%% file: main.tex
\pdfoutput=1

\documentclass[11pt]{article}

\usepackage{natbib}
\setcitestyle{numbers,square}
\usepackage{CUSTOM}

\usepackage{times}
\usepackage{latexsym}
\usepackage[T1]{fontenc}
\usepackage[utf8]{inputenc}
\usepackage{microtype}
\usepackage{inconsolata}
\usepackage{algorithm}
\usepackage{algorithmic}
\usepackage{siunitx}
\usepackage{amsmath}
\usepackage{graphicx}
\usepackage{multirow}
\usepackage{subcaption}
\usepackage{booktabs}
\usepackage{hyperref}
\usepackage{array} 
\usepackage{makecell}
\usepackage{enumitem}
\usepackage{longtable}
\usepackage{supertabular}
\usepackage{amssymb}
\usepackage{float}
\usepackage[capitalize]{cleveref}

\crefname{section}{Sec.}{Secs.}
\Crefname{section}{Section}{Sections}
\Crefname{table}{Table}{Tables}
\crefname{table}{Tab.}{Tabs.}

\graphicspath{ {./figure} }

\title{SRNN: Spatiotemporal Relational Neural Network for Intuitive Physics Understanding}

\author{Fei Yang \\
         CogBeauty Lab \\
         yftadyz@163.com}

\begin{document}
\maketitle
\begin{abstract}

Human prowess in intuitive physics remains unmatched by machines. To bridge this gap, we argue for a fundamental shift towards brain-inspired computational principles. This paper introduces the Spatiotemporal Relational Neural Network (SRNN), a model that establishes a unified neural representation for object attributes, relations, and timeline, with computations governed by a Hebbian ``Fire Together, Wire Together'' mechanism across dedicated \textit{What} and \textit{How} pathways. This unified representation is directly used to generate structured linguistic descriptions of the visual scene, bridging perception and language within a shared neural substrate. On the CLEVRER benchmark, SRNN achieves competitive performance, thereby confirming its capability to represent essential spatiotemporal relations from the visual stream. Cognitive ablation analysis further reveals a benchmark bias, outlining a path for a more holistic evaluation. Finally, the white-box nature of SRNN enables precise pinpointing of error root causes. Our work provides a proof-of-concept that confirms the viability of translating key principles of biological intelligence into engineered systems for intuitive physics understanding in constrained environments.

\end{abstract}

\input{1intro}
\input{2related}
\input{3method}
\input{4experiment}

\section{Conclusion}

In this work, we propose the Spatiotemporal Relational Neural Network (SRNN), which unifies object attributes, relations, and timeline through a Hebbian learning mechanism across dedicated pathways. Experimental results on CLEVRER demonstrate that principles of biological intelligence can be successfully engineered to understand intuitive physics. Through cognitive ablation, we outline a path toward a more holistic evaluation by identifying and mitigating benchmark bias. Evolving its core set of relations into a richer repertoire following the human developmental path will enable SRNN to interpret human activities in daily life. 

\bibliographystyle{plain}
\bibliography{custom}

\input{6appendix}

\end{document}

%% file: 1intro.tex
\section{Introduction}

Human beings possess a remarkable ability to understand and interact with the physical world, underpinned by a core component of human intelligence known as intuitive physics \cite{spelke2007core, lake2016buildingmachineslearnthink}. This innate capability allows us to form rich mental representations of objects and their spatiotemporal relationships. Even in infancy, humans demonstrate a sophisticated grasp of intuitive physics: they can discern object boundaries from visual input, mentally reconstruct the complete shapes of objects that become partially or fully occluded, and accurately anticipate the trajectories and final resting places of moving objects \cite{valenza2006perceptual}. 

The computer vision community has made significant progress in enabling machines to acquire this capability \citep{Qwen2.5-VL, wu2024deepseekvl2mixtureofexpertsvisionlanguagemodels, damonlpsg2025videollama3, actiongenome2020, Damen2018EPICKITCHENS}. However, their performance on visual cognitive understanding tasks still pales in comparison to human-level competence \citep{li2025core, 11094230}. To bridge this gap, we propose that a fundamental shift is necessary: we must look to the brain—the only proven general visual cognitive system—as an unparalleled blueprint. Our research, therefore, aims to identify the brain's core computational principles and implement them in artificial neural networks, thereby translating this biological success into engineered general intelligence.

Our model, the Spatiotemporal Relational Neural Network (SRNN), establishes a unified spatiotemporal representation in which object attributes (shape/texture/color), relations (touch, distance change, direction change, and kinematic states), and time are directly encoded as neurons. The core computation of forming this representation follows the Hebbian principle, often summarized as \textit{Fire Together, Wire Together} \citep{hebb1949organization}. This mechanism, mediated by stamp neurons, operates concurrently across three key components: a \textit{How} pathway for object relations, a \textit{What} pathway for object attributes, and a temporal binding process for timeline construction. Furthermore, SRNN extends this unified representation to bridge perception with language. Prevailing methods in AI often connect video and language by projecting both into a shared embedding space \citep{radford2021learning}. In contrast, the human brain seamlessly links visual perception and language through a network of interconnected neurons. Inspired by this architecture, SRNN directly grounds language generation in the same neural substrate that represents spatiotemporal concepts.

We evaluate SRNN on the CLEVRER \citep{CLEVRER2020ICLR} benchmark, a dataset comprising 20,000 synthetic videos of object collisions designed to probe physical reasoning. In our experiments, SRNN first generates a structured textual description for each video. This description, along with a question, is then processed by a large language model (LLM) to derive the final answer. Experimental results demonstrate that SRNN achieves competitive accuracy on CLEVRER, confirming its ability to effectively capture and encode essential spatiotemporal relations from visual stream. Moreover, a cognitive ablation study reveals a significant bias in the CLEVRER benchmark: its question distribution does not comprehensively cover the fundamental elements of human physical cognition. To address this gap, we outline a pathway—illustrated with concrete examples—toward constructing a more holistic benchmark for evaluating machines on human-like physical cognition. Finally, the white-box nature of SRNN enables us to precisely pinpoint the root cause of errors in every sampled case, revealing a diverse set of failure modes—from relational misidentification to LLM errors—that are often obscured in end-to-end black-box models. Code and models will be made publicly available soon.

%% file: 2related.tex
\section{Relative Work}

\begin{figure*}[t]
  \includegraphics[width=1\textwidth]{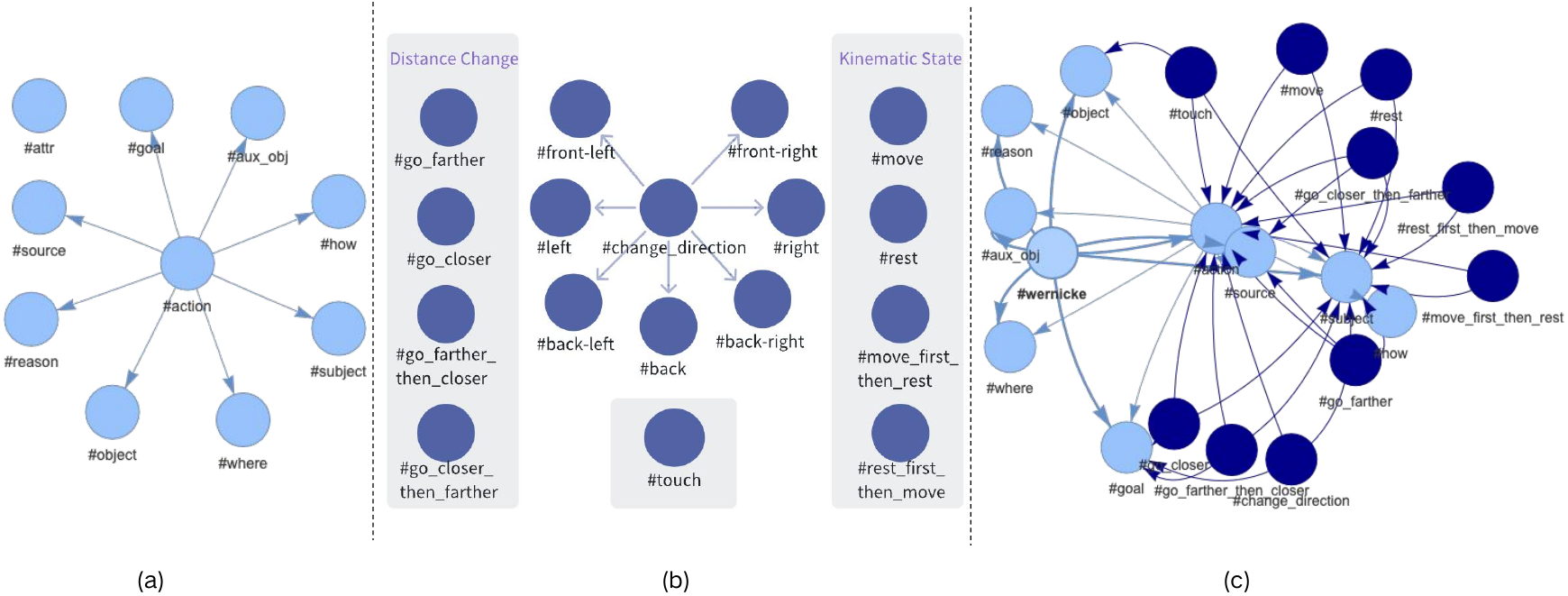}
  \caption{(a) Semantic Nature Design. The core of this design is \texttt{\#action} which connects eight semantic roles. (b) Spatial Nature Design. The spatial neurons represent spatial relations between objects, in addition to their intrinsic states. For \texttt{\#change\_direction}, we add \texttt{\#relation\_attr} neurons (e.g. \texttt{\#right}, \texttt{\#left}, \texttt{\#back}, etc.) to indicate its attributes. For other relations, \texttt{\#relation\_attr} does not exist.  (c) Connections between spatial neurons and semantic neurons. The neuron \texttt{\#wernicke} makes the semantic neurons capable of being activated when language is necessary. All neurons are depicted in blue to indicate their inactive state.}
  \label{fig:LoadNatureDesign}
\end{figure*}

\textbf{Intuitive Physics.} This topic refers to the implicit, informal expectations and understandings about the functioning of the physical world that humans possess prior to formal education \citep{spelke1994initial, baillargeon2007development, carey2011origin}. Cognitive science provides experimental evidence that humans, even in infancy, can comprehend touch \citep{leslie1987do}, distance change \citep{slater1990size}, direction change \citep{spelke1992origins}, and kinematic states \citep{kotovsky1998reasoning}. For example, infants show surprise when an object spontaneously begins moving from rest, as this violates the expectation that external force is required to initiate movement from a stationary state \citep{baillargeon1987object}. Machines must first comprehend these basic relationships that serve as the cornerstones of human cognition.

\textbf{Visual Processing.} Neurological evidence suggests that the human visual system decomposes the incoming spatiotemporal stream of information into two distinct pathways \citep{goodale1992separate, kravitz2011new}. The ventral stream (What Pathway) processes static object properties like color and shape, which are selectively attended to and bound to a coherent object representation through attentional mechanisms \citep{TREISMAN198097}. The dorsal stream (How Pathway) processes spatial location and motion. 
This information is processed in a hierarchical manner, from low-level features to complex representations \citep{felleman1991distributed}. Crucially, the brain shapes its complex representations via a process where frequently co-activated neurons strengthen their connections—a process often summarized as "neurons that fire together, wire together" \citep{hebb1949organization}. When describing the visual scene, these integrated signals are routed to language networks for interpretation, where conceptual semantics are processed in Wernicke's area and structured sentences are generated via Broca's area \citep{hickok2007cortical}. The architecture of our SRNN embodies the above neurological principles of human dynamic scene understanding, representing a functional, computational neuro-mimetic approach. 

\textbf{Intuitive Physics Reasoning Task.}
Neuro-symbolic approaches have been widely explored for this task. NS-DR~\cite{yi2019clevrer} represents the spatiotemporal relations of each frame in a video as a directed graph. However, a unified spatiotemporal graph is absent, as NS-DR does not have nodes to represent time or static attributes of an object. Moreover, unlike human brain, where relational concepts are stored in neuronal assemblies \citep{rizzolatti2004mirror}, NS-DR uses edges to represent these concepts. DCL~\cite{chen2021grounding} and VRDP~\cite{NEURIPS2021_07845cd9} represent the spatiotemporal relations in each frame of a video as a couple of matrix where each element stands for a numerical confidence score. Object attribute (e.g. shape, color) and relational concepts (e.g. collision) are represented as embedding vectors, thereby being able to optimize a goal function through backpropagation. In terms of representation and computation, the above methods are much different from human brain.

For end-to-end approaches, Object-based Attention~\cite{ding2020object} designs a transformer-based neural network for spatio-temporal reasoning about videos. All components are updated by self-supervised learning. In \citep{Lerer2016learning}, convolutional neural networks are trained to understand the physical intuition presented in block tower videos. Although the architecture of these neural networks have similarities with human brain, the computational mechanisms underlying neuron connection weights are fundamentally distinct from the Hebbian principle. Backpropagation conveys global error signals, whereas Hebbian rule relies exclusively on local activity.

Among existing benchmarks for intuitive physical reasoning \citep{CLEVRER2020ICLR, chen2025compositionalphysicalreasoningobjects, girdhar2020cater, riochet2018intphys, bordes2025intphys2benchmarkingintuitive, Bear2021Physion, tung2023physionplusplus}, we evaluate SRNN on CLEVRER \citep{CLEVRER2020ICLR}. This dataset contains 15,000 five-second videos showing object interactions, with objects defined by three static attributes—shape, texture, and color—and undergoing linear motion without magnetic forces. CLEVRER includes four types of questions: (1) predictive, requiring forecasting of future relations; (2) counterfactual, exploring alternative scenarios by removing objects; (3) descriptive, involving identification of current attributes and relationships; and (4) explanatory, demanding cause identification for observed relations. The collision events in CLEVRER are interpreted as touch relations in this work. We exclude other benchmarks due to elements uncommon in daily human scenes or visual complexity beyond our research scope. For instance, ComPhy \citep{chen2025compositionalphysicalreasoningobjects} introduces charge-based attraction and repulsion, while CATER \citep{girdhar2020cater} involves object levitation.

%% file: 3method.tex
\section{SRNN}

SRNN first boots spatial and semantic nature designs to predefine its neural network architecture. After that, it begins to watch a video. For each frame of the video, it perceives static states of objects. The attributes of shape, texture and color are attended. In the meanwhile, pixel coordinates of object centers are transformed into 3D camera coordinates by depth estimation. We track these camera coordinates to form object trajectories. Then it perceives spatiotemporal relations between each pair of objects given their trajectories within a time slot. After that, neurons are fired and wired, which occurs simultaneously within three key components: a \textit{How} pathway for object relations, a \textit{What} pathway for object attributes, and a temporal binding for timeline construction. A semantic network is also activated and generates sentences to describe the video. 

\subsection{Loading Nature Design}

SRNN initializes its cognitive architecture by loading a predefined neural network called Nature Design. We adopt a core distinction between the propensity that humans inherit since they are born (Nature Design) and knowledge acquired from developmental experiences (Nurture Belief) \citep{yang2024automatic}. The Nature Design mainly includes two parts: spatial neurons and semantic neurons, which are presented in Fig.~\ref{fig:LoadNatureDesign} (a) and (b), respectively. The semantic part draws inspiration from PropNet architecture as described in \citep{yang2025propnetwhiteboxhumanlikenetwork}. Fig.~\ref{fig:LoadNatureDesign} (c) shows the connections between spatial neurons and semantic neurons. Before watching a video, all these neurons are in an inactive state.

\subsection{Perceiving Static States}

We mimic human attention mechanism to attend to and integrate multiple visual attributes into coherent object percepts.
First when SRNN receives the frame data of a video, it obtains detection results for shape and texture of objects using YOLO-based models. Then cross-modal binding is performed by calculating the Intersection-over-Union (IoU) between shape and texture bounding boxes. Only detection pairs exceeding an IoU threshold are considered. Each region’s dominant color is extracted by aggregating RGB values. When shape and texture boxes overlap significantly and share the same color perception, their features are merged into a unified object representation with a composite label (e.g., “blue\_metal\_sphere”).

To support estimating the distance between objects in 3D space, we convert frame image pixels into egocentric camera coordinates. For each pixel at position (u, v), the depth value is computed by an opensource tool named \texttt{Video Depth Anything} \citep{Chen2025VideoDA}. The focal length of the camera is estimated by \texttt{GeoCalib} \citep{veicht2024geocalib}. The process of computing camera coordinates (x, y, z) is described by:

\begin{equation}
    \begin{aligned}
        x &= \frac{(u - c_x) \cdot z}{f} \\
        y &= \frac{(v - c_y) \cdot z}{f} \\
        z &= d(u, v)
    \end{aligned}
\end{equation}

where $z = d(u, v)$ is the depth (distance from the camera plane) at pixel $(u, v)$ and $f$ is the focal length (in pixels) of the camera. The $c_x, c_y$ are the coordinates of the principal point (the optical center of the image). We perform 3D scene reconstruction by back-projecting all pixels into a point cloud using \texttt{MeshLab}. The metric consistency and geometric plausibility of the derived egocentric coordinates are confirmed. Finally, the Euclidean distance between object centers is computed for each pair of objects appearing in the frame given their camera coordinates.

\subsection{Perceiving Spatiotemporal Relations}

Object tracking is performed based on bounding box overlap and class consistency across consecutive frames. 
Detections that fail to match any stored trace initiate a new object entity.

We split the video into non-overlapping time slots. Each time slot contains an equal number of consecutive frames.
Within each slot, object relations are computed from the object trajectories and the distance arrays between objects.
In this work, we focus on four relations: the kinematic profile of an object, direction change of an object, distance change between two objects, and touch between two objects. 

\textbf{Kinematic Profile of an Object}. The kinematic state of an object—whether it is moving or at rest—is determined by analyzing its trajectory over a sequence of frames. Specifically, we calculate the Euclidean distance between the center positions in the first and the last frame. If this net displacement exceeds a threshold parameter, the object is classified as \texttt{\#move}; otherwise it is classified as \texttt{\#rest}. For more granularity, the trajectory of \texttt{\#move} object is split into two equal-length segments to detect transitional states \texttt{\#rest\_first\_then\_move} or \texttt{\#move\_first\_then\_rest}.

\textbf{Direction Change of an Object}. Direction change is determined by comparing the principal movement vectors of an object’s trajectory before and after a touch. The signed angle between these vectors is computed, and a direction change occurs if the magnitude of the angle surpasses a threshold. Seven directions are predefined: \texttt{front-right}, \texttt{right}, \texttt{back-right}, \texttt{back}, \texttt{back-left}, \texttt{left}, \texttt{front-left}. The angular variation range for each direction is provided in App.~\ref{app:direction_labels}.

\textbf{Distance Change Between Two Objects}. A distance change occurs once the variation amplitude of the distance array exceeds a predefined threshold. Then we identify trend patterns by analyzing variations in differences between adjacent elements in the distance array. Four trends are defined: \texttt{\#go\_closer}, \texttt{\#go\_farther}, \texttt{\#go\_farther\_then\_closer}, and \texttt{\#go\_closer\_then\_farther}. If the minimum distance between the two objects exceeds a given threshold, the trend is not attended. As the time slot is very shot and object jittering does not exist, no other distance trends remain.

\begin{figure*}[t]
  \includegraphics[width=1\textwidth]{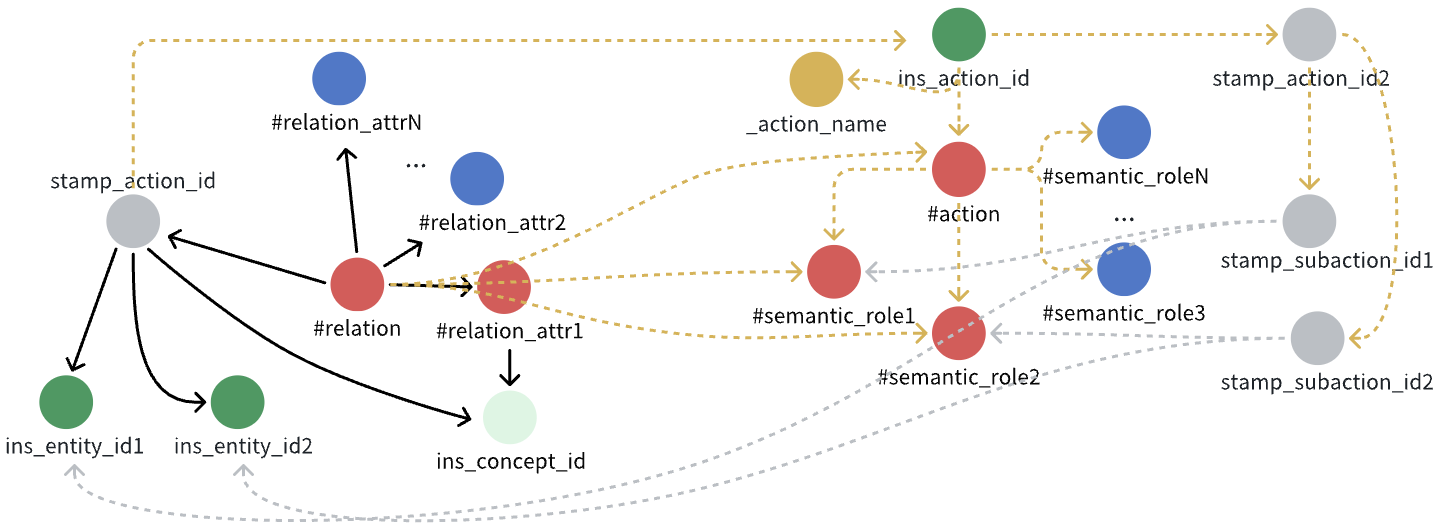}
  \caption{The Fire-and-Wire Mechanism (left) and the Language Generation Module (right) in How Pathway. \textbf{Left:} Visual perceptions activate a relational neuron \texttt{\#relation} along with creation and activation of entity-instance neurons \texttt{ins\_entity\_id1} and \texttt{ins\_entity\_id2}. Then \texttt{\#relation} triggers the creation and activation of a action-stamp neuron \texttt{stamp\_action\_id} which binds \texttt{ins\_entity\_id1} and \texttt{ins\_entity\_id2}. If relational attributes exists for this relation, the corresponding \texttt{\#relation\_attr1} is activated by \texttt{\#relation}. A concept-instance neuron \texttt{ins\_concept\_id1} is pointed by \texttt{\#relation\_attr1} and \texttt{stamp\_action\_id}. Other neurons below the activation threshold remain inactive (shown in blue).  
  \textbf{Right:} The \texttt{stamp\_action\_id} neuron triggers \texttt{ins\_action\_id}, which acts as the starting point of the semantic network. Connections are formed between \texttt{ins\_action\_id} and its lexical neuron \texttt{\_action\_name}, as well as with \texttt{\#action}. Meanwhile, \texttt{\#relation} propagates signals to \texttt{\#action} and \texttt{\#semantic\_roles} via predefined neural pathways in Nature Design. Joint signals from \texttt{ins\_action\_id} and \texttt{\#relation} activate \texttt{\#action}, which in turn sends signals to all the semantic-role neurons. Only \texttt{\#semantic\_role1} and \texttt{\#semantic\_role2} are activated upon receiving sufficient signals. Finally, \texttt{stamp\_subaction\_id1} wires \texttt{\#semantic\_role1} and \texttt{ins\_entity\_id1}, and \texttt{stamp\_subaction\_id2} binds \texttt{\#semantic\_role2} and \texttt{ins\_entity\_id2}.}
  \label{fig:fire_wire_how}
\end{figure*}

\textbf{Touch Between Two Objects}. When the inter-object distance falls below a touch threshold, auxiliary motions are further examined to confirm touch. At least one object in the pair under inspection must either exhibit a change in moving direction or transition from a stationary to moving state. To enhance reliability, we also enforce an isolation condition: neither object should be close to a third object below the touch threshold within the time slot, reducing false positives in crowded scenes. Since the time slot is extremely short, this condition does not noticeably reduce the recall of touches.

The threshold parameters for each relation are determined in a human-driven fashion to ensure alignment with human perception. Specifically, we calibrate these parameters on a handful of videos and then validate the computational results against human judgments. The tuning process is illustrated in App.~\ref{app:model_parameters}.

\subsection{Firing and Wiring Neurons}

Inspired by the neurobiological concept that ``synchronous firing strengthens synaptic connections'' \citep{hebb1949organization},
we designed a representation mechanism where SRNN processes the perceived static states and spatiotemporal relations by activating relevant neurons (\textit{Fire}) and then establishing connections between these activated neurons (\textit{Wire}). 
This process encodes both how entities interact or change over time and what attributes of entities are present in the neural network.

Each neuron possesses a threshold and becomes activated when its cumulative input within a time slot reaches this threshold, subsequently emitting a signal. Signals propagate along directed connections to other neurons. All neurons emit signals with a value of 1.

\textbf{How Pathway.} 

Neurons (e.g., \texttt{\#move}, \texttt{\#touch}) in Spatial Nature Design are activated by the observed corresponding relations. 
All participant instances involved in these relations are represented by entity-instance neurons which simultaneously receive activation.
When a relation neuron fires, it triggers the creation of an action-stamp neuron, which connects all the participating entities of the relation.
The left network of Fig.~\ref{fig:fire_wire_how} details this process.

\begin{figure}[t]
  \includegraphics[width=0.9\columnwidth]{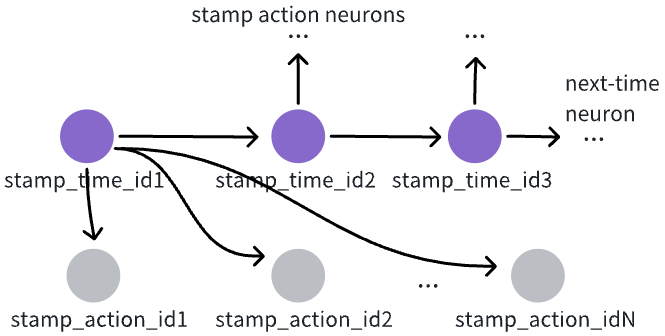}
  \centering
  \caption{The Fire-and-Wire Mechanism in Temporal Binding. The temporal neuron \texttt{stamp\_time\_id1} marks the origin of the timeline and points to \texttt{stamp\_time\_id2}. All action-stamp neurons present during this temporal window are associated with \texttt{stamp\_time\_id1}. As time progresses, new temporal neurons are generated, activated, and linked sequentially to form an ordered chain.}
  \label{fig:fire_wire_time}
\end{figure}

\textbf{Temporal Binding.} 
A unique time-stamp neuron is generated for each time slot, sequentially linked to the previous time-stamp neuron. 
This neuron connects all action-stamp neurons within the same time slot, binding co-occurring relations into a coherent episodic memory trace.
The details of this process are depicted in Fig.~\ref{fig:fire_wire_time}.

\textbf{What Pathway.}
For each entity detected in the relations, its attended attributes like color and texture are represented by concept-instance neurons which are linked by \texttt{\#attr}.
When the entity neuron fires, it triggers the creation of an entity-stamp neuron, which connects all the attribute neurons of the entity.
Fig.~\ref{fig:fire_wire_what} illustrates the process details.

\subsection{Generating Language Description}
\label{sec:gen_lang_desc}
\begin{figure}[t]
  \includegraphics[width=0.9\columnwidth]{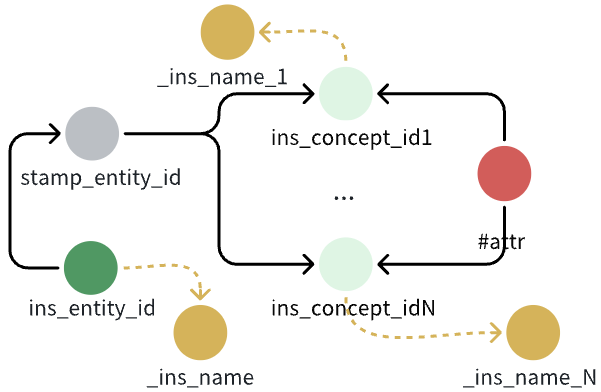}
  \centering
  \caption{The Fire-and-Wire Mechanism and the Language Generation Module in What Pathway. The neuron \texttt{ins\_entity\_id} initiates the creation and activation of \texttt{stamp\_entity\_id}. Simultaneously, concept-instance neurons are created and activated to encode the entity's attributes, which are then bound together by \texttt{stamp\_entity\_id}. These instance neurons, in turn, activate corresponding lexical neurons (indicated by yellow dashed lines), denoted as \texttt{\_ins\_name}.}
  \label{fig:fire_wire_what}
\end{figure}

The language module generates a sentence description for each relation. 
It contains two components: a semantic network and a sentence generator.
The semantic network fulfills three primary functions: (1) assigning lexical labels to objects; (2) assigning verbal labels to relations; and (3) assigning semantic roles like agent and patient to the participants involved in a relation.
In this project, the moving object in a relation serves as the agent. 
We present this semantic network in detail in Fig.~\ref{fig:fire_wire_how} and Fig.~\ref{fig:fire_wire_what} with yellow dashed lines.
The implementation of the semantic network draws inspiration from the approach outlined in \citep{yang2025propnetwhiteboxhumanlikenetwork}. 
In contrast to their framework, a key distinction lies in our introduction of subaction-stamp neurons to bind semantic roles and instance entities, thereby explicitly defining the relationship between entities and their corresponding roles.

The sentence generator takes the semantic network as input and generates a sentence. 
A sentence structure is sequentially constructed by positioning the agent entity first, followed by the relation lemma, patient entity (if present), and goal or source entities introduced by the prepositions ``to'' or ``from'' respectively.
Entity attributes are incorporated by querying associated concept neurons to enrich descriptions with conceptual features (e.g., color, texture). 
The final sentence is generated by concatenating all components into a fluent string. 
For example, ``the rubber gray sphere ins\_entity\_39 touch rubber yellow sphere ins\_entity\_37.'' 

\subsection{Predicting Future Relations}
To make SRNN be able to handle counterfactual and predictive questions from CLEVRER, 
we add this module to predict if an object will touch another object in the next time slot or after the video ends.
This algorithm employs PCA \citep{jolliffe2016principal} to identify the principal direction of motion from 3D trajectory data. 
It projects the trajectory onto this axis and applies linear regression to derive a constant-velocity model, assuming zero acceleration for straight-line motion prediction.
Predictions are discarded for entity pairs initially beyond a distance threshold to ensure physically plausible forecasts.
Predicted results are appended in text form following the output of Section~\ref{sec:gen_lang_desc}.

In App.~\ref{app:example}, we provide an example that details the output of the aforementioned modules.

%% file: 4experiment.tex
\section{Experiments}

We first demonstrate SRNN achieves competitive accuracy on the CLEVRER dataset, proving its effectiveness in capturing essential video information. 
Next, a cognitive ablation study is conducted to quantify the causal role of specific relations and timeline in video reasoning. This study also serves to audit inherent biases in CLEVRER for the development of a more comprehensive benchmark. Finally, a fine-grained error analysis is performed on incorrectly answered cases, leveraging the intrinsic interpretability of our model to categorize failure patterns and identify limitations.

\input{tables/main_comparison}

\subsection{Performance Comparison}
\label{exp:main_comparison}

We compare the performance of SRNN on CLEVRER against a comprehensive set of baselines, including: TVQA+~\cite{lei2018tvqa}, Memory~\cite{fan2019heterogeneous}, IEP (V)~\cite{johnson2017inferring}, TbD-net (V)~\cite{mascharka2018transparency}, HCRN~\cite{le2020hierarchical}, MAC (V+)~\cite{hudson2018compositional}, NS-DR~\cite{yi2019clevrer}, DCL~\cite{chen2021grounding}, Object-based Attention~\cite{ding2020object}, and VRDP~\cite{NEURIPS2021_07845cd9}. Among these, VRDP~\cite{NEURIPS2021_07845cd9} represents the state-of-the-art neuro-symbolic model, while Object-based Attention~\cite{ding2020object} is the state-of-the-art end-to-end approach.

\textbf{Evaluation Metrics.} We employ question answering accuracy as the evaluation metric. It is important to note that for the multi-choice questions (including explanatory, predictive, and counterfactual types), a question is considered correctly answered (per ques.) only if all its associated options (per opt.) are answered correctly.

\textbf{Implementation Details.} Each video is temporally segmented into five time slots, with each slot corresponding to a one-second duration. The video descriptions generated by SRNN are utilized as input for a Large Language Model (LLM) to answer the questions. Specifically, for a given video and a question, the textual descriptions from SRNN are concatenated into a prompt along with the question text. To ensure consistent comprehension between humans and LLM, critical alignments of the question contexts are incorporated into the prompt. Furthermore, specific solving instructions are added to guide LLM in reasoning through the problem based on the provided logic. All parameters and prompt alignments are tuned on ten videos from the CLEVRER training split, and the final performance is evaluated on the CLEVRER validation split. The results of key parameters are presented in App.~\ref{app:model_parameters}. YOLO training details are appended in App.~\ref{app:yolo_models}. We select \texttt{Qwen3}~\citep{qwen3} for this task, and more details about the LLM choice and its prompts are provided in App.~\ref{app:prompt}.

\textbf{Results.} As shown in Tab.~\ref{tab:main_comparison}, our method delivers a strong performance of 90.5\% per option and 80.0\% per question for explanatory questions. This result places it as the third-best performing model in our comparison, after Object-based Attention (98.5\% per opt., 96.0\% per ques.) and VRDP (96.3\% per opt., 91.9\% per ques.). On descriptive questions, our method achieves 89.2\% accuracy, a result that is on par with the top-performing models, slightly behind VRDP (93.4\%).
For counterfactual questions, our method attains 84.7\% per option and 58.0\% per question, approaching Object-based Attention (91.4\% per opt., 75.6\% per ques.). These results collectively affirm SRNN's effectiveness in capturing and encoding the essential relational information presented in a video, which is necessary for complex spatio-temporal and causal reasoning.

\input{tables/cognitive_ablation}
\subsection{Cognitive Ablation}
\label{exp:cognitive_ablation}

In this section, we perform a cognitive ablation study. 
Unlike conventional ablation that removes structural components of a model (e.g., layers or neurons), 
we remove the human-understandable physical concepts (e.g., touch, distance change) and the timeline presented in the videos, to directly interrogate the causal role of specific relations in video reasoning.
Moreover, we audit the biases of the CLEVRER dataset to provide a roadmap for designing a more comprehensive benchmark.

By selectively removing relational neurons in the neural network, such as \texttt{\#touch}, corresponding physical concepts can be effectively stripped from the video. 
For timeline ablation, we randomize the linking order of the time-stamp neurons to simulate the temporal disorder of relational events in the video.
The resulting textual descriptions are then fed into LLM to answer questions and evaluate accuracy. 
The configurations of SRNN and LLM remain identical to those used in Sec.~\ref{exp:main_comparison}.
Tab.~\ref{tab:cognitive_ablation} presents the changes in answer accuracy of the model when different types of relations are removed for various question categories.

\textbf{Predictive Questions.} 
The drastic performance drop from Time Order ablation (-25.3\%/-46.1\%) demonstrates that forecasting is critically dependent on a precise understanding of relation sequences. 
Furthermore, the significant impact of Future Touch ablation (-9.5\%/-16.7\%) confirms that the accuracy is directly driven by the capacity to anticipate imminent collisions.
These results reveal that the predictive questions entirely fail to address stationary objects, alterations in object orientation, and variations in inter-object distances presented in the video.

\textbf{Counterfactual Questions.} 
The profound performance degradation resulting from Touch ablation (–15.9\%/–35.3\%) confirms that tangible collisions serve as the primary anchor for mental simulation. 
Concurrently, the notable impact of Time Order ablation (–7.1\%/–16.0\%) underscores that constructing coherent counterfactuals also requires altering the sequence of relations. 
These findings expose a pronounced bias in the CLEVRER dataset, where the counterfactual questions heavily rely on inverting or modifying contact-based interactions. 
To address this bias, the set of answer choices could be expanded to incorporate options based on other factors, 
such as kinematic state (e.g., ``The yellow rubber object stays at rest after the blue cylinder is removed'') 
or changes in relative distance (e.g., ``The yellow rubber object goes close to the red cube sphere after the blue cylinder is removed'').

\textbf{Descriptive Questions.} 
The severe performance degradation caused by Touch ablation (-15.2\%) establishes that this type of questions places a significant emphasis on the touch interactions in the video.
Concurrently, the substantial drop from Time Order ablation (-18.9\%) confirms that accurately describing object attributes and relations often necessitates an implicit understanding of the temporal context in which they occur.
CLEVRER largely overlooks the roles of Direction Change and Distance Change. 
Future benchmarks should be extended to probe these under-represented concepts,
such as "How many objects are moving away from the red metal sphere when the rubber object exits the scene" (Distance Change)
or "Which direction does the yellow rubber object shift to after it touches the green cube" (Direction Change).

Descriptive questions also target static objects, 
such as ``How many metal objects are stationary when the blue rubber object exits the scene'' or 
``What color is the stationary rubber object''. 
The significant accuracy drop (-4.8\%) directly confirms their important role in this task.
In contrast, other types of questions show almost no performance drop after removing this factor, 
indicating that they do not rely on these static objects that serve as background. 
This also reveals a pronounced bias in the CLEVRER dataset.

\textbf{Explanatory Questions.} 
The severe accuracy drop from Time Order ablation (-27.9\%/-49.1\%) proves that causal attribution depends almost entirely on timing. This aligns with expectations, as explaining an event necessarily relies on its past temporal sequence. The major decline under Touch ablation (-25.2\%/-40.8\%) confirms that explanation prioritizes the touch relations in the video. This exposes a strong bias in CLEVRER, where the explanatory questions overwhelmingly focus on collision. A more balanced benchmark should incorporate questions involving non-contact influences, such as ``Which of the following is responsible for the brown object's going close to the purple object'' (Distance Change) or "Which of the following is responsible for the brown object's changing direction to its left?" (Direction Change).

By treating the LLM as a fixed reasoning engine, the drastic performance drop when removing specific relations proves that the LLM's success is not inherent but is critically enabled by the information within our representation. Therefore, this study cleanly separates the SRNN's role as the information provider from the LLM's role as a powerful reasoner.

\input{tables/error_distribution}
\subsection{Error Analysis}

We conduct a fine-grained error analysis to identify the limitations of SRNN on the CLEVRER dataset and to guide future improvements. For each question type, we randomly sample 20 incorrectly answered cases from Sec.~\ref{exp:main_comparison}, ensuring that each multi-choice problem appears only once. We then categorize the failure patterns and analyze their distribution across different question types.

A key advantage of our approach lies in its case-level interpretability. Unlike embedding methods that rely on opaque numerical representations, our model allows us to inspect individual neurons at each time slot via an interactive visualization interface. This capability enables us to precisely pinpoint the root cause of errors in every sampled case, revealing a diverse set of error types that would be difficult to discern in embedding models. Error types are defined as follows:
\begin{itemize}[noitemsep, topsep=4pt, leftmargin=1.2em]
    \item \textbf{Ambiguous Ground Truth}. Human verifiers cannot reliably determine the correctness of the ground-truth answers in the benchmark.
    \item \textbf{Annotation Error}. The ground-truth answers for questions in the benchmark are incorrect.
    \item \textbf{Relation Timing Error}. Incorrect identification of the timing of relations, like misidentification of the sequence of two touches.
    \item \textbf{False Positive Touch}. SRNN anticipates a touch relation that fails to occur in the video or after the video.
    \item \textbf{Friction Neglect}. Neglect of friction effects, leading to errors in the module of future touch prediction. This module assumes uniform linear motion only.
    \item \textbf{LLM Description Misunderstanding}. LLM misunderstands the video description text.
    \item \textbf{LLM Incomplete Prompt}. Incomplete reasoning logic in the prompt, preventing the LLM from outputting correct answers.
    \item \textbf{LLM Prompt Deviation}. LLM does not strictly follow the prompt's guidance during problem-solving.
    \item \textbf{Missed Touch}. Failure to detect touch relations. It occurs when a collision is extremely minor, or when the object's trajectory remains unaltered after a collision.
    \item \textbf{Object Detection Failure}. Failure to detect objects in some frames. This error primarily occurs when the YOLO models fail to detect objects or produce inaccurate detections.
    \item \textbf{Occluded Collision Path}. The path to collision is blocked by a third object, and this factor is not considered by our algorithm.
    \item \textbf{Out Of Attention}. Objects are too distant, so the future prediction module is not activated.
    \item \textbf{Question Misalignment}. A discrepancy exists between the intended meaning of the question and the LLM's interpretation of it, leading to reasoning based on a misunderstood premise.

\end{itemize}

Based on the error distribution summarized in Tab.~\ref{tab:error_distribution}, we analyze the primary error types to identify the key bottlenecks of SRNN across different question categories.

\textbf{Predictive Questions.} 
The analysis reveals that the future prediction module is a primary bottleneck, accounting for approximately 50\% of all errors. This is directly evidenced by the failure modes: Out Of Attention (20\%), Occluded Collision Path (10\%), Missed Touch (10\%), and Friction Neglect (10\%), which provides concrete directions for optimizing the future prediction component.

\textbf{Counterfactual Questions.} 
The primary bottleneck is Missed Touch (55\%), where it fails to detect critical touch mentioned in questions. This error deprives the LLM of necessary contextual information, preventing correct answer generation.

\textbf{Descriptive Questions.} 
The main sources of error include Object Detection Failure (20\%), Question Misalignment (15\%), and Ambiguous Ground Truth (15\%). As this question type fundamentally relies on accurately describing scene content, failures in object recognition have an immediate and pronounced impact. Additionally, the diverse phrasing of descriptive questions introduces greater challenges in achieving semantic alignment between the LLM's interpretation and the question's intent.

\textbf{Explanatory Questions.} 
The primary challenge involves touch relation recognition, with Missed Touch (35\%) and False Positive Touch (15\%) constituting half of all errors. Both missing critical contacts and hallucinating non-existent ones corrupt the causal chain, depriving the LLM of the correct physical context required for generating valid explanations.

%% file: tables/main_comparison.tex
\begin{table*}[t]
	\begin{center}
	\small
	\setlength{\tabcolsep}{5pt}
	\resizebox{1\linewidth}{!}{%
	\begin{tabular}{lccccccc}
	\toprule
        \multirow{2}{*}{Methods} & \multicolumn{2}{c}{Predictive} & \multicolumn{2}{c}{Counterfactual} & \multirow{2}{*}{Descriptive} & \multicolumn{2}{c}{Explanatory} \\ 
    \cmidrule(lr){2-3}\cmidrule(lr){4-5}\cmidrule(lr){7-8}
        & per opt. & per ques. & per opt. & per ques. &  & per opt. & per ques. \\ 
    \midrule
        \midrule
        TVQA+~\cite{lei2018tvqa} & 70.3 & 48.9 & 53.9 & 4.1 & 72.0 & 63.3 & 23.7\\    
        Memory~\cite{fan2019heterogeneous}  & 50.0 & 33.1 & 54.2& 7.0 & 54.7 & 53.7 & 13.9 \\
        IEP (V)~\cite{johnson2017inferring} & 50.0 & 9.7 & 53.4 & 3.8 & 52.8 & 52.6 & 14.5  \\ 
        TbD-net (V)~\cite{mascharka2018transparency}  & 50.3 & 6.5 & 56.1 & 4.4 & 79.5 & 61.6 & 3.8 \\
        HCRN~\cite{le2020hierarchical} & 54.1 & 21.0 & 57.1 & 11.5 & 55.7 & 63.3 & 21.0 \\ 
        MAC (V+)~\cite{hudson2018compositional} & 59.7 & 42.9 & 63.5 & 25.1 & 86.4 & 70.5 & 22.3 \\ 
        NS-DR~\cite{yi2019clevrer}  & 82.9 & 68.7 & 74.1 & 42.2 & 88.1 & 87.6 & 79.6 \\
        DCL~\cite{chen2021grounding} & 90.5 & 82.0 & 80.4 & 46.5 & 90.7 & 89.6 & 82.8 \\
        Object-based Attention~\cite{ding2020object} & 93.5 & 87.5 & 91.4 & 75.6 & \textbf{94.0} & \textbf{98.5} & \textbf{96.0} \\
        VRDP~\cite{NEURIPS2021_07845cd9}  & \textbf{95.7} & \textbf{91.4} & \textbf{94.8} & \textbf{84.3} & 93.4 & 96.3 & 91.9 \\
        \midrule
        SRNN + LLM & 81.0 & 62.9 & 84.7 & 58.0 & 89.2 & 90.5 & 80.0 \\
    \bottomrule
	\end{tabular}}
	\end{center}
	\caption{Comparison of question-answering accuracy on CLEVRER. The highest accuracy for each question type is highlighted in boldface. All baseline scores are sourced directly from \cite{NEURIPS2021_07845cd9}. SRNN demonstrates notably competitive accuracy.}
	\label{tab:main_comparison}
	\vspace{-10pt}
\end{table*}

%% file: tables/cognitive_ablation.tex
\begin{table*}[t]
	\begin{center}
	\small
	\setlength{\tabcolsep}{5pt}
	\resizebox{1\linewidth}{!}{%
	\begin{tabular}{lccccccc}
	\toprule
        \multirow{2}{*}{Ablation Dimension} & \multicolumn{2}{c}{Predictive} & \multicolumn{2}{c}{Counterfactual} & \multirow{2}{*}{Descriptive} & \multicolumn{2}{c}{Explanatory} \\ 
    \cmidrule(lr){2-3}\cmidrule(lr){4-5}\cmidrule(lr){7-8}
        & per opt. & per ques. & per opt. & per ques. &  & per opt. & per ques. \\ 
    \midrule
        -- & 81.0 & 62.9 & 84.7 & 58.0 & 89.2 & 90.5 & 80.0 \\
        Rest State & -1.8 & -2.9 & -0.1 & -0.5 & -4.8 & -0.7 & -0.2 \\
        Direction Change & -1.1 & -2.2 & -0.6 & -2.7 & -0.2 & -0.4 & -0.1 \\
        Distance Change & -0.3 & -1.3 & -0.3 & -1.5 & 0.1 & -0.3 & -1.3 \\
        Touch & -2.9 & -4.2 & -15.9 & -35.3 & -15.2 & -25.2 & -40.8 \\
        Future Touch & -9.5 & -16.7 & -4.7 & -12.6 & 0.2 & -0.6 & -0.5 \\
        Time Order & -25.3 & -46.1 & -7.1 & -16.0 & -18.9 & -27.9 & -49.1 \\
    \bottomrule
	\end{tabular}}
	\end{center}
	\caption{The first row presents the baseline accuracy, while the following rows display the decrease in accuracy upon the removal of each dimension. \textit{Rest State} implies complete disregard for stationary objects.}
	\label{tab:cognitive_ablation}
	\vspace{-10pt}
\end{table*}

%% file: tables/error_distribution.tex
\begin{table}[t]
\centering
\caption{Error type distribution (in \%) across predictive (P), counterfactual (CF), descriptive (D), and explanatory (E) questions. The top three errors for each question type are highlighted in boldface.}
\label{tab:error_distribution}
\begin{tabular}{l *{4}{S[table-format=2.0]}}
\toprule
\textbf{Error Type} & \textbf{P} & \textbf{CF} & \textbf{D} & \textbf{E} \\
\midrule
Ambiguous Ground Truth & \textbf{20} & \textbf{15} & \textbf{15} & 0 \\
Annotation Error & 0 & 5 & 10 & 5 \\
Relation Timing Error & 0 & 0 & 10 & 10 \\
False Positive Touch & 0 & 5 & 10 & \textbf{15} \\
Friction Neglect & 10 & 0 & 0 & 0 \\
LLM Misunderstanding & 5 & 5 & 10 & 5 \\
LLM Incomplete Prompt & 0 & 0 & 0 & \textbf{15} \\
LLM Prompt Deviation & 10 & 0 & 0 & 10 \\
Missed Touch & 10 & \textbf{55} & 10 & \textbf{35} \\
Object Detection Failure & \textbf{15} & 0 & \textbf{20} & 5 \\
Occluded Collision Path & 10 & 0 & 0 & 0 \\
Out Of Attention & \textbf{20} & \textbf{15} & 0 & 0 \\
Question Misalignment & 0 & 0 & \textbf{15} & 0 \\
\bottomrule
\end{tabular}
\end{table}

%% file: 6appendix.tex
\clearpage
\appendix
\section{YOLO Models}
\label{app:yolo_models}

We annotated shape and texture bounding boxes for 190 randomly selected frames from CLEVRER training videos using LabelImg, a Python package for image annotation. The frame set is distributed as follows: 25\% contain boundary objects, 25\% contain obstructed objects, and the remaining 50\% contain only normal objects (without boundary or obstruction). The data was split into 160 training and 30 validation images. Two separate YOLO11m models were trained for shape and texture detection with the following parameters: shape model (lr=0.002, epochs=20, batch=4) and texture model (lr=0.005, epochs=20, batch=4). Both models achieved near-perfect detection performance across all categories, which is presented in Tab.~\ref{tab:yolo_performance}.

\begin{figure}[h]
  \raggedleft
  \includegraphics[width=0.95\columnwidth]{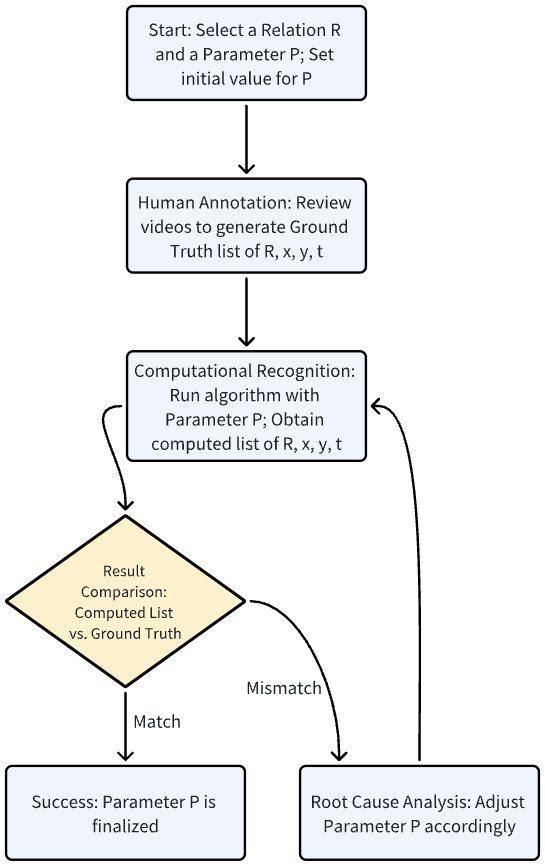}
  \caption{Human-Driven Parameter Tuning Loop. In the list notation, entities $x$ and $y$ participate in relation $R$ at time slot $t$.}
  \label{fig:model_param_tuning_loop}
\end{figure}

\begin{table}[h]
\centering
\caption{Detection performance of shape and texture models}
\begin{tabular}{lccc}
\hline
Model & Class & Precision & Recall \\
\hline
\multirow{3}{*}{Shape}
& Sphere & 1.000 & 0.996 \\
& Cube & 0.994 & 1.000 \\
& Cylinder & 0.986 & 1.000 \\
\hline
\multirow{2}{*}{Texture}
& Rubber & 0.999 & 1.000 \\
& Metal & 0.999 & 1.000 \\
\hline
\end{tabular}
\label{tab:yolo_performance}
\end{table}

\begin{table}[h]
\centering
\caption{Direction Labels and Corresponding Angle Intervals}
\label{tab:direction_intervals}
\begin{tabular}{lc}
\toprule
\textbf{Direction} & \textbf{Angle Interval (degrees)} \\
\midrule
Front & $(-22.5, 22.5]$ \\
Front-right & $(22.5, 67.5]$ \\
Right & $(67.5, 112.5]$ \\
Back-right & $(112.5, 157.5]$ \\
Back (positive) & $(157.5, 180]$ \\
Back (negative) & $(-180, -157.5]$ \\
Back-left & $(-157.5, -112.5]$ \\
Left & $(-112.5, -67.5]$ \\
Front-left & $(-67.5, -22.5]$ \\
\bottomrule
\end{tabular}
\end{table}

\input{supps/supps_parameters}
\section{Model Parameters}
\label{app:model_parameters}
The threshold parameters for each relation are determined in a human-driven fashion, guided by human perceptual judgments to ensure consistency. We calibrate these parameters using 10 videos from the CLEVRER training set and validate the computing result of the relation against human observations, with the detailed process shown in Fig.~\ref{fig:model_param_tuning_loop}. Tab.~\ref{tab:parameters} presents core parameter definitions and their values after tuning.

\section{Direction Labels}
\label{app:direction_labels}

Tab.~\ref{tab:direction_intervals} shows the angle interval for each direction label. The back direction is split into two separate intervals—Back (positive) and Back (negative)—due to the circular nature of angular measurements. In a standard angular coordinate system, angles wrap around at ±180°, meaning that -180° and +180° represent the same physical direction (directly behind). This discontinuity necessitates dividing the back region into two intervals to maintain mathematical consistency while covering the complete 45° angular span centered at 180° (or equivalently, -180°). The positive interval $(157.5, 180]$ covers angles approaching from the right side, while the negative interval $(-180, -157.5]$ covers angles approaching from the left side, together forming a continuous back-facing sector.

\section{SRNN Examples}
\label{app:example}

\begin{figure*}[t]
  \centering
  \includegraphics[width=0.95\textwidth]{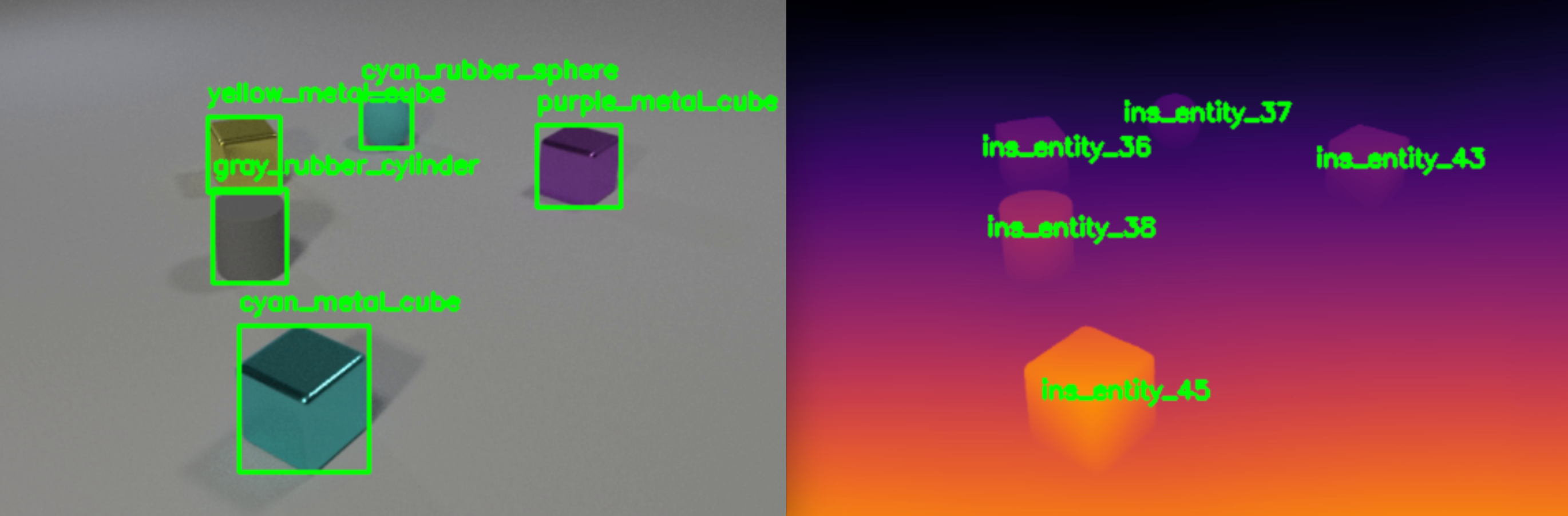}
  \caption{Object detection (left) and depth estimation (right) results in a single CLEVRER video frame. The left panel shows detected object regions—shape and texture—bound into unified representations (e.g., ``blue metal sphere'') through IoU-based fusion and color categorization. The right panel presents the corresponding depth map, where each pixel’s depth is estimated using Video Depth Anything.}
  \label{fig:supps_video_snapshot}
\end{figure*}

\begin{figure*}[t]
  \includegraphics[width=1\textwidth]{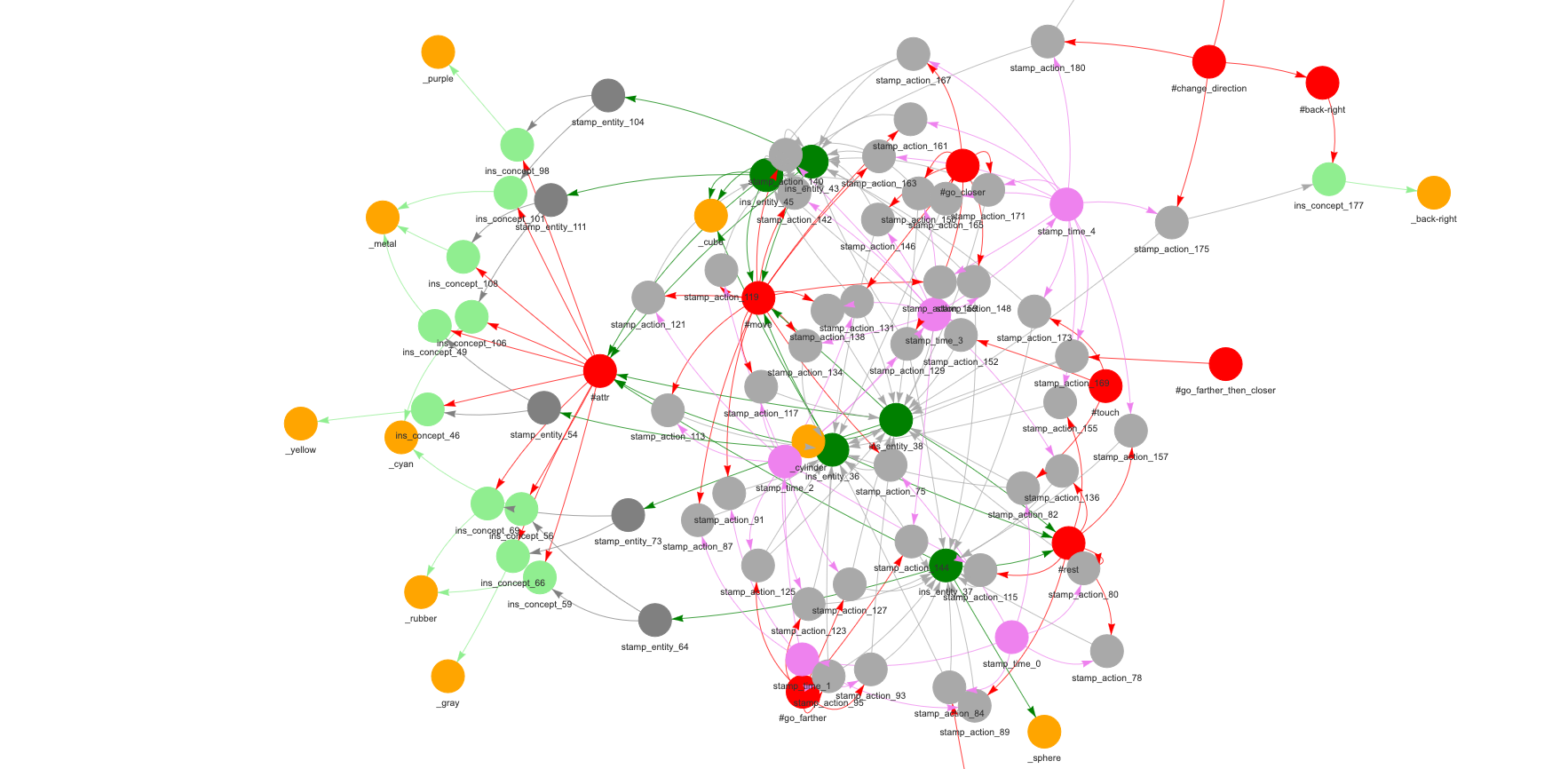}
  \caption{Neural network representation for \texttt{video\_12214} presented in an interactive visualization interface. In this representation, neurons are color-coded according to their type: nature neurons are shown in red, stamp neurons in gray, time neurons in purple, instance neurons in green, and lexical neurons in yellow. Deactivated neurons are omitted for clarity. Arrows on the connections indicate the direction of signal transmission upon neuron activation. Neurons can be dragged to visualize their connections with other neurons.}
  \label{fig:supps_drnn}
\end{figure*}

\begin{figure*}[h]
  \includegraphics[width=0.95\textwidth]{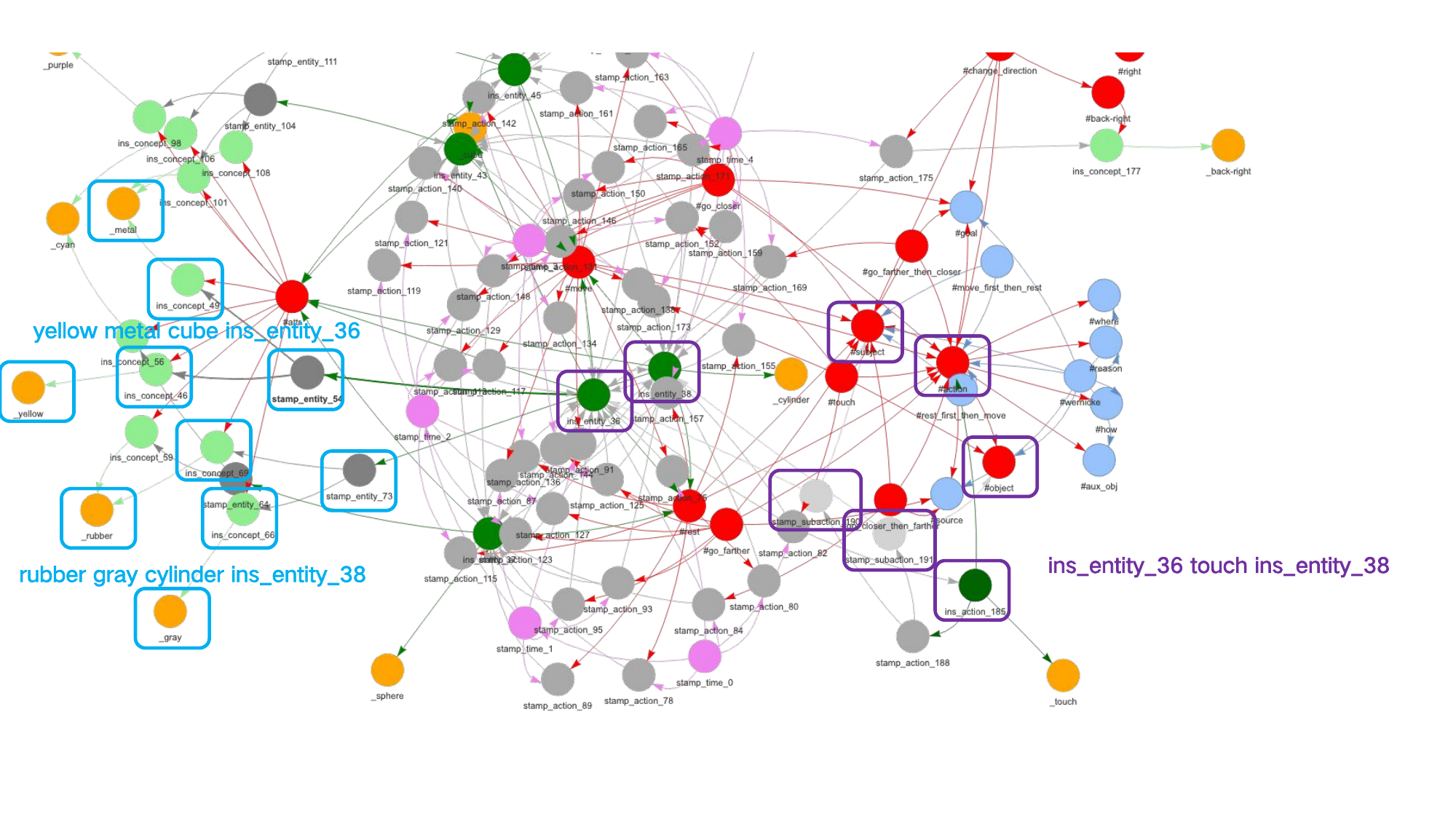}
  \caption{Language generation module for \texttt{video\_12214}. \textbf{Right}: The purple boxes on the right mark the language generation process for the How Pathway. A neuron named \texttt{ins\_action\_185} is created and activated by the neural network, serving as the start point of the semantic network. A connection is established between ins\_action\_185 and \texttt{\#action} by the neural network after the birth of \texttt{ins\_action\_185}, making \texttt{ins\_action\_185} being able to send a signal to \texttt{\#action}. The neural network also establishes a connection between \texttt{ins\_action\_185} and a lexical neuron \texttt{\_touch}. Next, two semantic role neurons \texttt{\#subject} and \texttt{\#object} are activated by the joint signals from \texttt{\#touch} and \texttt{\#action}. Finally, \texttt{stamp\_subaction\_190} is created to bind \texttt{ins\_entity\_36} and \texttt{\#subject}. Similarly, \texttt{stamp\_subaction\_191} is created to bind \texttt{ins\_entity\_38} and \texttt{\#object}. The sentence generator outputs a sentence ``ins\_entity\_36 touch ins\_entity\_38'' based on this semantic network. \textbf{Left}: The left cyan boxes illustrate the language generation process for the What Pathway.
The neuron \texttt{ins\_entity\_36} triggers the creation and activation of \texttt{stamp\_entity\_54}. 
In the meantime, \texttt{ins\_concept\_46}, \texttt{ins\_concept\_49}, etc., are created and activated to record the attributes of this entity, and bound by \texttt{stamp\_entity\_54}. A phrase ``yellow metal cube ins\_entity\_36'' is generated based on this semantic network. Similarly, another phrase ``rubber gray cylinder ins\_entity\_38'' is also produced. The sentence generator then merges the outputs from the How Pathway and the What Pathway, ultimately yielding the final sentence: ``The yellow metal cube ins\_entity\_36 touch the rubber gray cylinder ins\_entity\_38.'' Note that the language generation module currently does not include handling of subject-verb agreement.}
  \label{fig:supps_language_final}
\end{figure*}

\begin{figure*}[t]
  \includegraphics[width=0.95\textwidth]{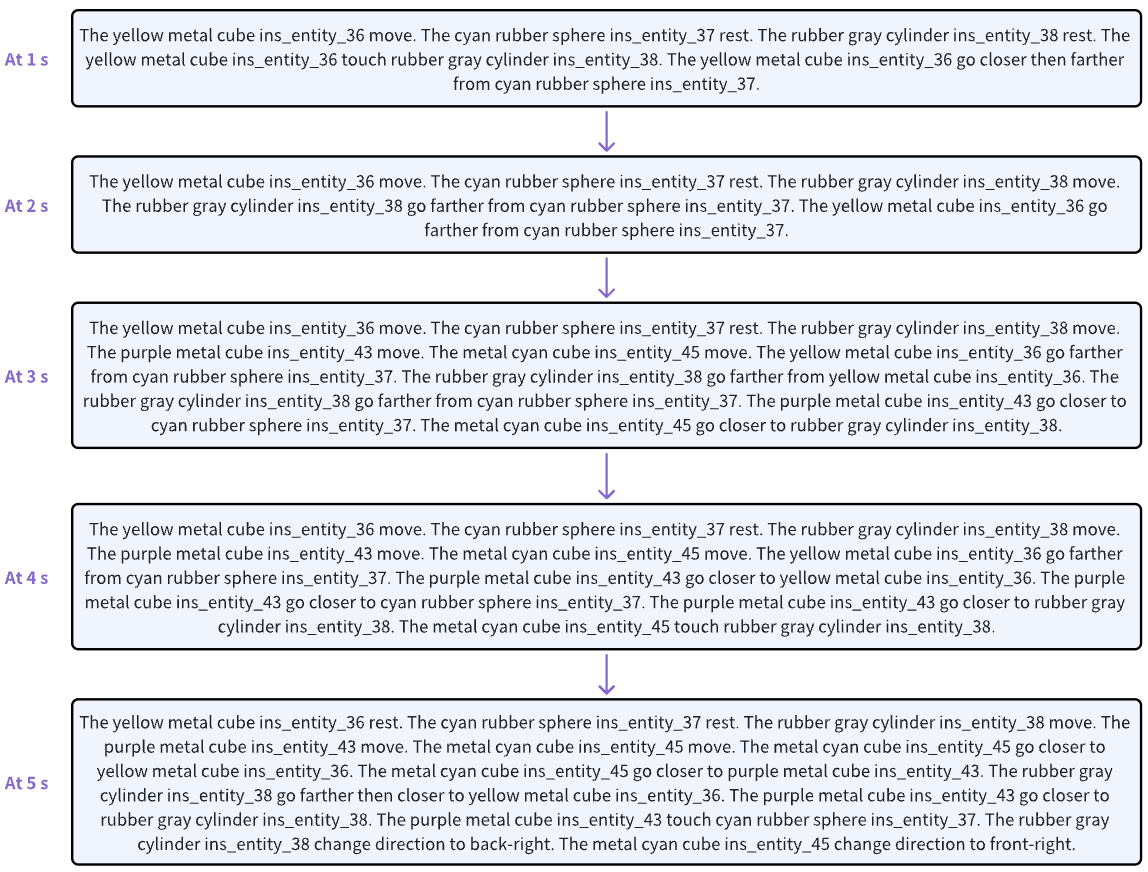}
  \caption{Description text for \texttt{video\_12214}. Each segment contains multiple sentences, with each describing a relation depicted in the video.}
  \label{fig:supps_video_desc_text}
\end{figure*}

Fig.~\ref{fig:supps_video_snapshot} presents an example of our approach for perceiving static properties of objects in a single frame from \texttt{video\_12214}. 
Fig.~\ref{fig:supps_drnn} illustrates the neural network  representation corresponding to \texttt{video\_12214}. Fig.~\ref{fig:supps_language_final} presents the language generation module for \texttt{video\_12214}.
Figure~\ref{fig:supps_video_desc_text} presents the textual descriptions for each time slot of \texttt{video\_12214}. 

\section{LLM Prompts}
\label{app:prompt}

We selected \texttt{Qwen3} as the LLM for CLEVRER question-answering tasks, with the model code \texttt{qwen3-235b-a22b-instruct-2507}. \texttt{Qwen3} is an open-source large language model that has demonstrated strong performance on reasoning benchmarks. In \cref{exp:main_comparison} and \cref{exp:cognitive_ablation}, we do not compare with other LLMs such as ChatGPT, Claude, Gemini, or DeepSeek, nor do we test smaller variants of \texttt{Qwen3}. First, a state-of-the-art LLM can be regarded as a “universal function approximator” for reasoning tasks. Its extensive knowledge and strong inference ability ensure that relevant information in the input is effectively leveraged, making further comparisons with such models superfluous. Second, using a smaller version of \texttt{Qwen3} would introduce a weaker reasoner. In that case, any failures could stem from the model’s own limitations rather than missing information in the representation, which would confound our analysis.

\begin{figure*}[t]
  \includegraphics[width=0.95\textwidth]{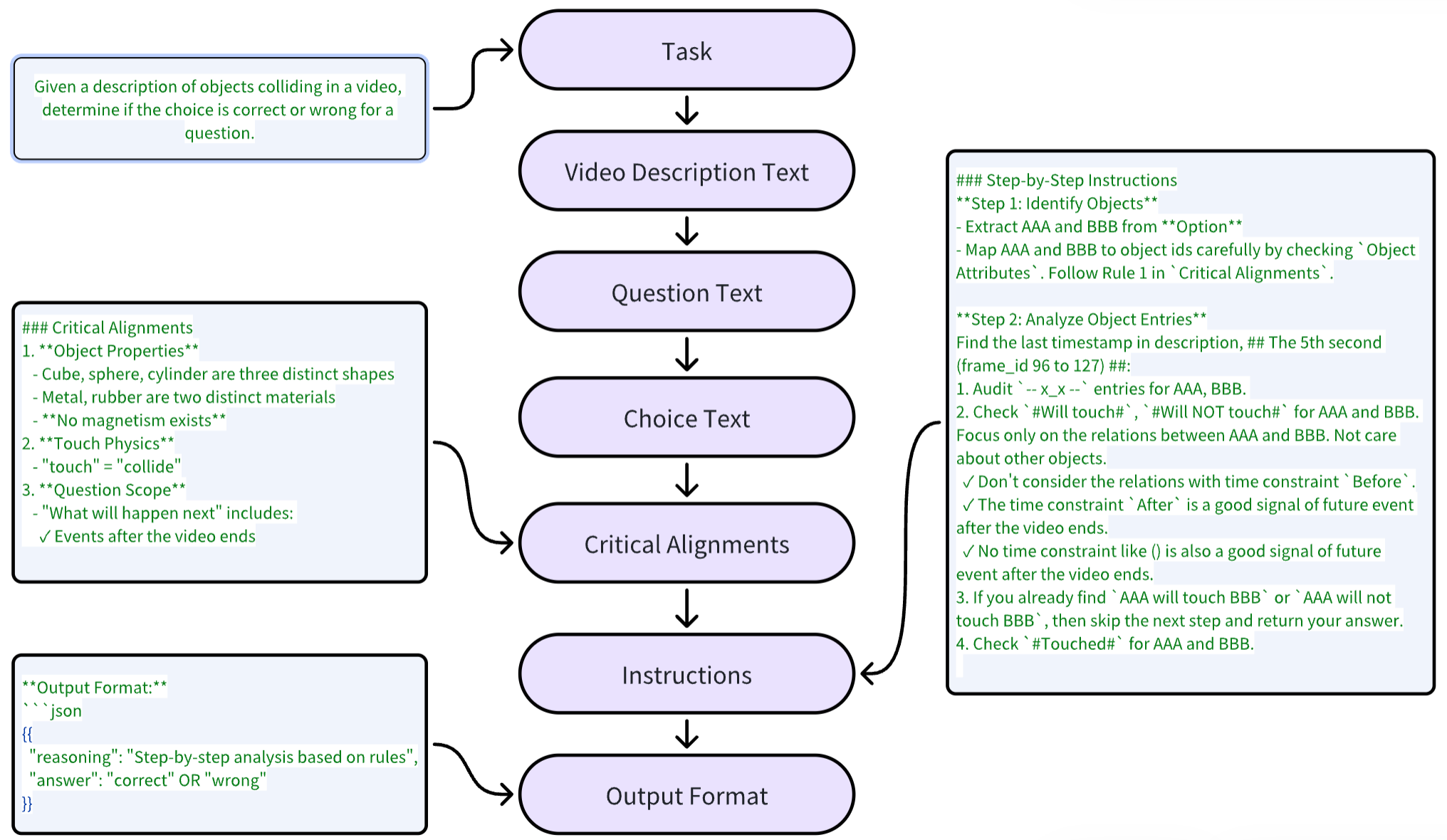}
  \caption{LLM prompt for predictive questions. The corresponding text segments are assembled into a prompt according to the arrow sequence.}
  \label{fig:supps_prompt_pred}
\end{figure*}

\begin{figure*}[t]
  \includegraphics[width=0.95\textwidth]{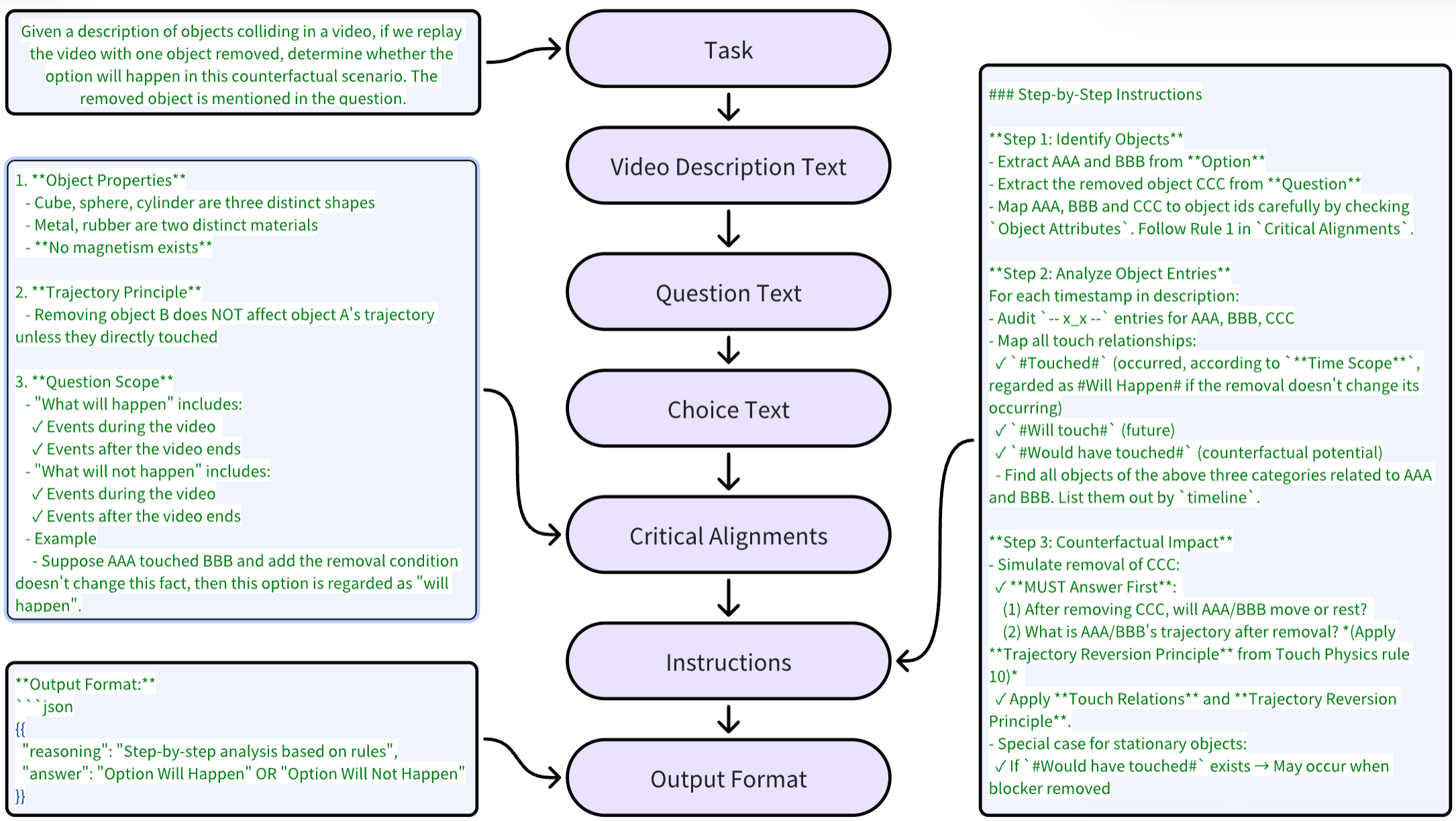}
  \caption{LLM prompt for counterfactual questions. The corresponding text segments are assembled into a prompt according to the arrow sequence.}
  \label{fig:supps_prompt_cf}
\end{figure*}

\begin{figure*}[t]
  \includegraphics[width=0.95\textwidth]{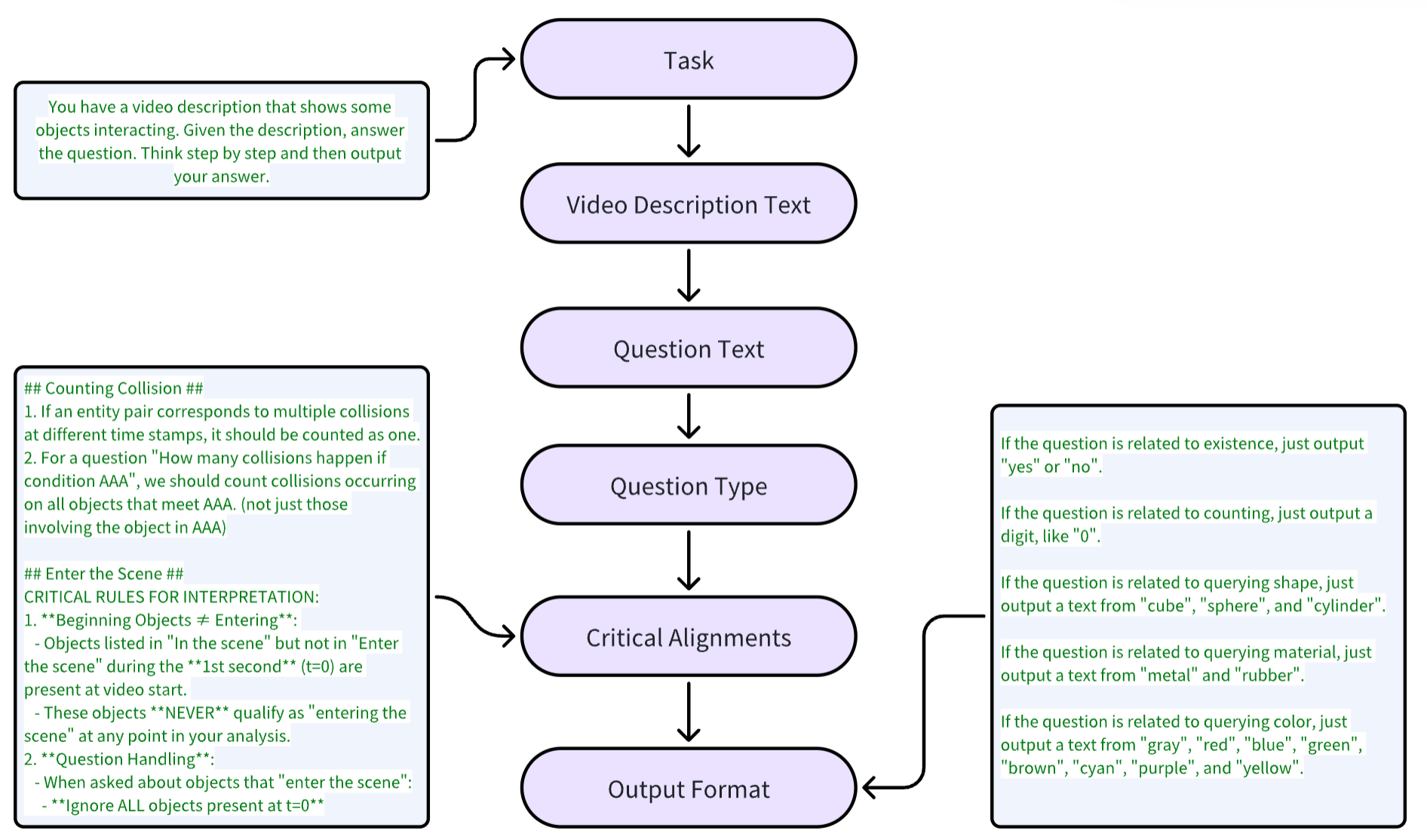}
  \caption{LLM prompt for descriptive questions. The question type of a descriptive question refers to existence, counting, querying shape, querying material or querying color. The corresponding text segments are assembled into a prompt according to the arrow sequence.}
  \label{fig:supps_prompt_desc}
\end{figure*}

\begin{figure*}[t]
  \includegraphics[width=0.95\textwidth]{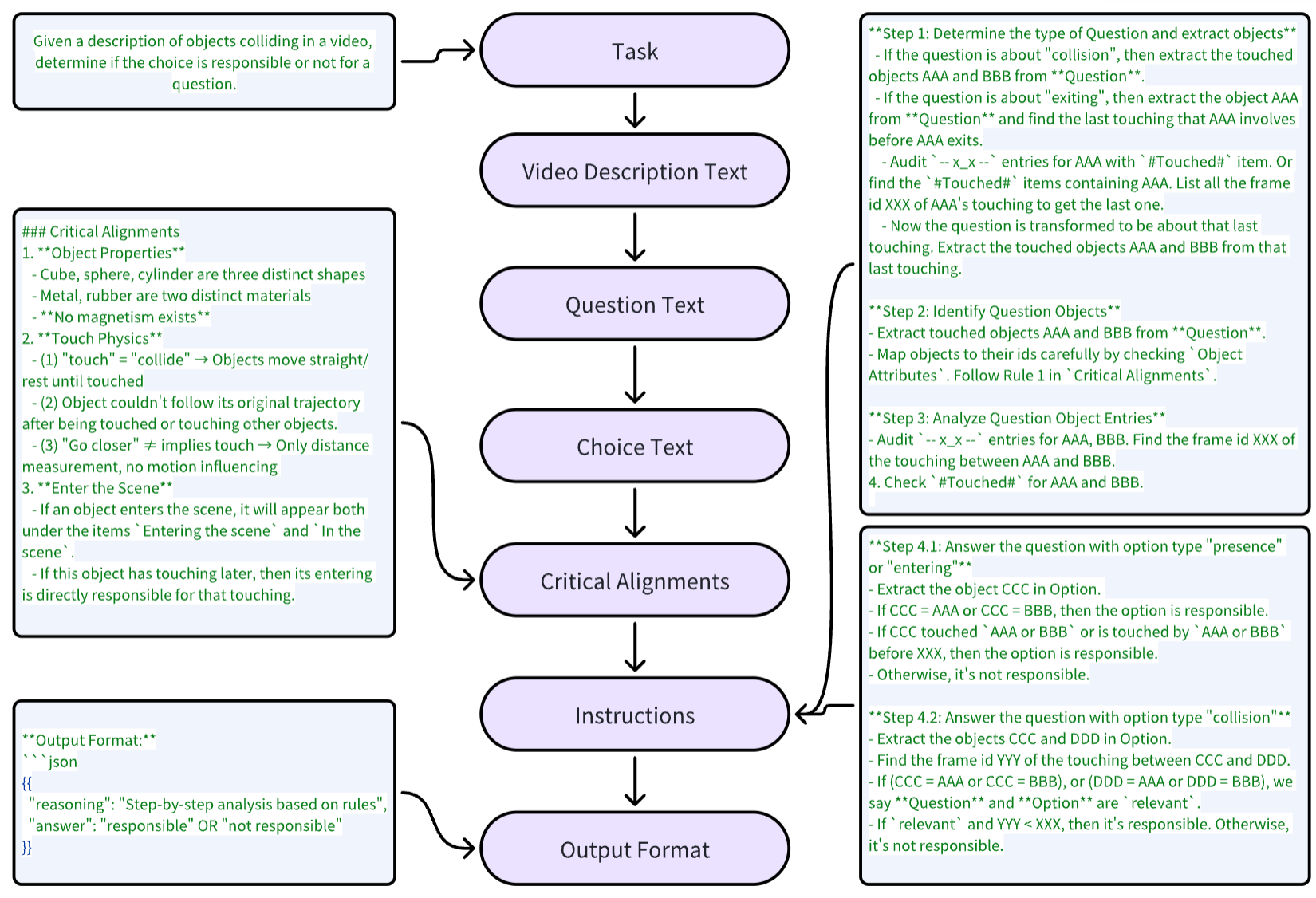}
  \caption{LLM prompt for explanatory questions. The corresponding text segments are assembled into a prompt according to the arrow sequence.}
  \label{fig:supps_prompt_expl}
\end{figure*}

The prompt comprises Task Definition, Video Description Text, Question Text, Choice Text, Critical Alignments, and Output Format. The prompts of predictive questions, counterfactual questions, descriptive questions, and explanatory questions are shown in Fig.~\ref{fig:supps_prompt_pred}, Fig.~\ref{fig:supps_prompt_cf}, Fig.~\ref{fig:supps_prompt_desc}, and Fig.~\ref{fig:supps_prompt_expl}, respectively.

LLM's understanding of the questions are not aligned with the question answers. Consider this question ``Which of the following is responsible for the gray sphere's colliding with the purple object''. LLM's understanding of this question is that only the most recent collision is responsible. However, the ground-truth answer is that all related collisions on the timeline are responsible.

Through iterative error analysis, we identify primary discrepancies between the LLM’s comprehension of questions and the corresponding ground-truth answers, which are subsequently integrated into the prompt.

%% file: supps/supps_parameters.tex
\begin{table*}[t]
\centering
\caption{SRNN Parameter Configurations}
\label{tab:parameters}
\renewcommand{\arraystretch}{1.5}
\begin{tabular}{p{0.35\linewidth}p{0.6\linewidth}}
\toprule
\textbf{Parameter Setting} & \textbf{Description} \\
\midrule
ATTENTION\_IOU\_THD = 0.85 & Bounding-box overlap threshold for attribute binding (IoU) \\
COLOR\_FOCUS\_AREA\_RATIO = 0.7 & Ratio of the central area used for color recognition \\
CONFIDENCE\_THD = 0.7 & Confidence threshold for object detection by YOLO models \\
FOCAL\_LENGTH = 420 & Used in transforming pixel coordinates to camera coordinates \\
BOX\_OVERLAP\_THD = 0.1 & Bounding-box overlap threshold for object tracking (IoU) \\
MOVE\_THD = 0.01 & Moving threshold for object kinematic state (Euclidean distance) \\
DIST\_ATT\_THD = 0.1 & If the minimum distance between two objects exceeds this threshold, the distance change is not attended \\
MOVING\_AVG\_WINDOW = 5 & Moving average window for computing distance change between two objects \\
TOUCH\_THD = 0.04 & Distance threshold for touch detection (Euclidean distance) \\
TOUCH\_BOX\_OVERLAP\_THD = 0.01 & Bounding-box overlap threshold for touch detection (IoU) \\
\bottomrule
\end{tabular}
\end{table*}